%% file: main.tex
\documentclass{bmvc2k}

\title{TSR-Ego: Temporally Guided Stereo Refinement Framework for Egocentric 3D Human Pose Estimation}

\addauthor{Md Mushfiqur Azam}{mdmushfiqur.azam@utsa.edu}{1}
\addauthor{John Quarles}{john.quarles@utsa.edu}{1}
\addauthor{Kevin Desai}{kevin.desai@utsa.edu}{1}

\addinstitution{
The University of Texas at San Antonio\\
}

\runninghead{Azam, Quarles, Desai}{TSR-Ego}

\usepackage[T1]{fontenc}
\usepackage{graphicx}
\usepackage{amsmath}
\usepackage{amssymb}
\usepackage[dvipsnames]{xcolor}
\usepackage{tikz}
\usetikzlibrary{calc}
\usetikzlibrary{positioning}
\usepackage{multirow}
\usepackage{booktabs}
\usepackage{subcaption}
\usepackage{wrapfig}
\usepackage{array}
\usepackage{capt-of}

\begin{document}

\maketitle

\input{sec/0_abstract}

\input{sec/1_intro}

\input{sec/2_related_works}

\input{sec/4_method}

\input{sec/5_Experiment}

\input{sec/6_conclusion}

\section{Acknowledgements}
This material is partially based upon work supported by the National Science Foundation under Grant No. 2316240 and 2403411. Any opinions, findings, and conclusions or recommendations expressed herein are those of the author(s) and do not reflect National Science Foundation views.

\bibliography{main}

\end{document}

%% file: sec/0_abstract.tex
\begin{abstract}
Egocentric 3D human pose estimation from head-mounted stereo cameras is challenging due to fisheye distortion, severe self-occlusion, and frequent truncation of body joints outside the camera field of view. Recent stereo egocentric methods have improved performance through heatmap lifting, stereo correspondence, and transformer-based refinement, but they often rely heavily on frame-local evidence or use temporal information only as auxiliary pose-level context. This limits robustness when current-frame stereo cues are weak, occluded, or ambiguous. We propose TSR-Ego, a temporally guided stereo framework that couples short-term motion evidence with projection-guided feature sampling. The model first enriches dense stereo feature maps using a causal depthwise-separable temporal convolution, allowing past visual evidence to influence the feature space before deformable cross-attention. A single-stage causal stereo decoder then refines learned 3D joint queries through temporal self-attention, joint self-attention, and fisheye deformable stereo cross-attention, using the evolving pose estimate to generate 2D sampling references. Unlike methods that apply temporal reasoning mainly after pose prediction, TSR-Ego uses motion context to shape both the sampled stereo features and the joint representations while preserving online inference without future frames. Experiments on UnrealEgo2 and UnrealEgo-RW show state-of-the-art performance, with especially strong gains on real-world sequences where single-frame stereo observations are unreliable. Code will be available at: \url{https://github.com/}.
\end{abstract}

%% file: sec/1_intro.tex
\section{Introduction}
\label{sec:intro}

Egocentric 3D human pose estimation~\cite{egocentricsurvey, unrealego,mo2cap2,selfpose, sceneego,invisiblepose, autocalib, egoego, ego3dpose} from head-mounted cameras is a key component for immersive AR/VR, telepresence, human-computer interaction, and embodied motion understanding. Unlike third-person pose estimation~\cite{deeppose,openpose,densepose}, the egocentric setting observes the body from an extreme first-person viewpoint, where fisheye distortion, large perspective variation, self-occlusion, and limited field of view make many joints only partially visible or completely out of view. These challenges are particularly severe for lower-body joints, which are often weakly observed by head-mounted cameras and must be inferred from incomplete visual evidence. Stereo fisheye cameras provide an attractive sensing setup because they introduce cross-view geometric cues while remaining compatible with compact wearable devices. However, effectively exploiting these cues remains difficult when body parts are truncated, stereo evidence is weak or one-sided, and visible joint evidence changes rapidly over time.

Recent datasets and methods have substantially advanced stereo egocentric 3D pose estimation. Large-scale benchmarks such as UnrealEgo, UnrealEgo2, and UnrealEgo-RW provide synthetic and real-world head-mounted fisheye stereo data for studying full-body pose estimation under severe occlusion and limited field of view~\cite{unrealego, 3dposeperception}. Building on these benchmarks, early approaches often rely on intermediate 2D heatmaps and lift them to 3D, which provides strong image-space localization but remains underconstrained when joints are occluded or outside the camera view. Recent stereo egocentric methods have improved 3D pose estimation by lifting image-space heatmaps into 3D poses, exploiting stereo correspondence and egocentric geometric cues, and refining joint representations with transformer-based attention~\cite{egoattention,ego3dpose,egoposeformer,3dposeperception}. 
Despite this progress, existing stereo egocentric methods still leave an important gap between spatial stereo reasoning and temporal guidance. Heatmap-based methods depend on intermediate 2D detections, geometry-aware methods rely on reliable current-frame stereo cues, and transformer-based refinement methods often build their 3D hypotheses primarily from the current stereo observation. Video-based approaches introduce temporal context, but temporal information is commonly used to augment joint representations or improve pose consistency rather than to condition the stereo features sampled during refinement. This separation is limiting in egocentric views, where joints may be occluded, truncated by the headset field of view, or visible in only one fisheye camera. In such cases, current-frame spatial evidence alone may produce an unreliable pose hypothesis, even though recent frames provide useful cues about joint motion and visibility.

We address this limitation with \textbf{TSR-Ego}, a temporally guided stereo refinement framework for egocentric 3D human pose estimation. Our key idea is to inject temporal evidence at the feature level, so that deformable stereo cross-attention samples from temporally enriched representations rather than only frame-local features. Given a causal window of stereo fisheye frames, TSR-Ego first applies a causal depthwise-separable convolution stack along the time axis to enrich each per-view spatial feature map with past visual evidence while preserving the spatial layout required for deformable sampling. A single-stage causal stereo decoder then refines learned 3D joint queries through interleaved causal temporal self-attention, joint self-attention, and fisheye deformable stereo cross-attention. The temporal self-attention updates each joint query using its own motion history, enabling online inference without future frames, while stereo sampling references are projected from the decoder's evolving pose estimate.

Unlike approaches that treat temporal reasoning as a post-processing or pose-level refinement step, TSR-Ego lets motion context condition both the stereo feature maps and the joint queries inside the decoder. This is useful in egocentric stereo settings where joints may be occluded, truncated, or visible in only one fisheye view, but remain predictable from recent motion. Our contributions are summarized as follows:
\begin{itemize}

\item A \textbf{Causal Temporal Feature Mixer (TFM)}, a causal depthwise-separable convolution over the time axis of stereo feature maps that enriches each spatial location with past visual evidence before deformable stereo cross-attention, conditioning the feature space itself rather than only the queries.

\item A single-stage causal stereo decoder that integrates pixel-level temporal feature enrichment with joint-local temporal reasoning for egocentric 3D human pose estimation.
\end{itemize}

We achieve state-of-the-art results on UnrealEgo2 and UnrealEgo-RW, validating the benefit of embedding temporal evidence inside stereo feature refinement rather than applying it only after pose prediction.

%% file: sec/2_related_works.tex
\section{Related Work}
\label{sec:related}
 Early monocular egocentric methods studied pose estimation from downward-facing head-mounted cameras, often relying on synthetic supervision or optimization-based temporal priors to handle missing body evidence~\cite{egoglobal, xregopose}. EgoPW~\cite{egopw} further moved toward in-the-wild egocentric pose estimation using weak external supervision, while SceneEgo~\cite{sceneego} introduced scene-aware constraints by projecting image and depth features into a voxel representation to improve physically plausible pose estimation during human-scene interaction. These works demonstrate the importance of geometric cues in egocentric pose estimation, but most monocular settings remain underconstrained when large body regions are occluded or outside the field of view.
\paragraph{Egocentric pose datasets and benchmarks.}
Egocentric human pose estimation has evolved from controlled capture setups toward large-scale, in-the-wild, and multimodal benchmarks. Early full-body egocentric datasets mainly used head-mounted monocular or stereo cameras to study the camera wearer's pose under severe viewpoint distortion, self-occlusion, and limited field of view~\cite{egocap, xregopose, mo2cap2}. Later datasets improved realism and supervision by introducing in-the-wild egocentric captures, weak external supervision, scene constraints, and human-interaction settings~\cite{egopw, sceneego}. More recently, large-scale egocentric resources such as Ego4D, Ego-Exo4D, and Nymeria have broadened the field from isolated pose estimation to first-person video understanding, paired ego-exo learning, multimodal sensing, and daily human motion modeling~\cite{ego4d, egoexo4d, nymeria}. However, many of these datasets are either not designed specifically for estimating the camera wearer's full-body 3D pose from stereo fisheye input, or rely on external views and additional sensors for supervision. In contrast, stereo fisheye benchmarks such as UnrealEgo, UnrealEgo2, and UnrealEgo-RW directly target head-mounted full-body pose estimation with complementary views and temporal video evidence~\cite{unrealego, 3dposeperception}. These benchmarks highlight the central challenge addressed in this work: exploiting both cross-view geometry and short-term temporal context to recover body joints under truncation, fast motion, and egocentric self-occlusion.

\paragraph{Stereo and geometry-aware egocentric pose estimation.}
Several stereo egocentric methods exploit intermediate 2D heatmaps, stereo correspondence, or perspective geometry to recover 3D pose. UnrealEgo~\cite{unrealego} uses stereo inputs with 2D keypoint estimation to improve 3D pose prediction, establishing the effectiveness of binocular fisheye observations for egocentric motion capture. Ego3DPose~\cite{ego3dpose} identifies two important cues in stereo egocentric views: cross-view correspondence and strong perspective variation of nearby limbs. It introduces a limb-wise two-path architecture and a perspective-aware orientation representation to better estimate 3D limb structure from binocular heatmaps. EgoTAP~\cite{egoattention} further improves heatmap-to-3D lifting by encoding stereo heatmaps with a Grid-ViT module and propagating information through a skeleton-aware network, allowing visible joints to support the estimation of occluded or weakly observed joints. These methods show that stereo geometry and skeletal structure are critical for egocentric pose estimation. However, their predictions still depend heavily on the quality of current-frame heatmaps or correspondence cues, which can be unreliable when joints are truncated, occluded, or visible in only one fisheye view.

\paragraph{Transformer and video-based egocentric pose estimation.}
Transformer architectures\cite{transformer} have improved 3D human pose estimation significantly in recent years. It enables flexible aggregation across spatial regions, joints, views, and time~\cite{uplift,exploitingtemporal, mhformer, towardsrobustpose, ustformer,stcformer, mixste, poseformer, poseformerv2, diff3dhpe, posegtac}. Recent egocentric methods have adapted transformer-based attention to better exploit spatial, temporal, and view-dependent cues. Ego-STAN~\cite{egostan} introduces domain-guided spatio-temporal self-attention for monocular fisheye egocentric pose estimation, using feature-map tokens to model fisheye distortion and self-occlusion. In the stereo setting, EgoPoseFormer~\cite{egoposeformer} formulates egocentric 3D pose estimation as a two-stage transformer framework, where a coarse 3D pose proposal is refined using deformable stereo attention over fine-grained multi-view features. EgoTAP~\cite{egoattention} also uses a ViT-style heatmap encoder before skeleton-aware lifting, indicating the benefit of attention-based feature modeling for stereo egocentric pose. More recently, Akada et al.~\cite{3dposeperception} incorporate depth-aware scene features and temporal context into a stereo-video framework, showing that scene geometry and motion history are useful for challenging motions such as sitting, crouching, and strong self-occlusion. These works show that attention and temporal context are effective for egocentric perception. However, temporal information is often used to enrich joint features, aggregate video representations, or improve pose consistency, while the stereo refinement process itself remains largely dependent on current-frame spatial evidence.

Our work builds on stereo egocentric pose estimation, transformer-based joint refinement, and temporal video reasoning. Unlike heatmap-lifting methods~\cite{egoattention,ego3dpose}, TSR-Ego predicts 3D poses directly without intermediate 2D detections. Compared with EgoPoseFormer~\cite{egoposeformer}, which refines a current-frame proposal, TSR-Ego injects causal temporal context into both stereo feature maps and joint queries before deformable cross-attention, enabling motion-aware stereo refinement.

%% file: sec/4_method.tex
\section{Method}
\label{sec:method}
Given a causal window of synchronized stereo observations
$\mathcal{I}_{t}=\{I_{\tau}^{v}\}_{\tau=t-T+1}^{t}$,
$v\in\{1,\ldots,V\}$, our model predicts a 3D pose sequence
$\hat{P}\in\mathbb{R}^{T\times J\times 3}$.

TSR-Ego uses a single-stage causal stereo decoder initialized with learned
per-joint queries and learned 3D reference points, avoiding an explicit
proposal stage. The decoder refines these queries through temporal reasoning,
body-structure reasoning, and geometry-aware stereo cross-attention. Temporal
evidence is introduced both at the pixel-feature level before stereo sampling
and at the query level after joint tokens are formed.

\begin{figure*}[t]
\centering
\includegraphics[width=\textwidth]{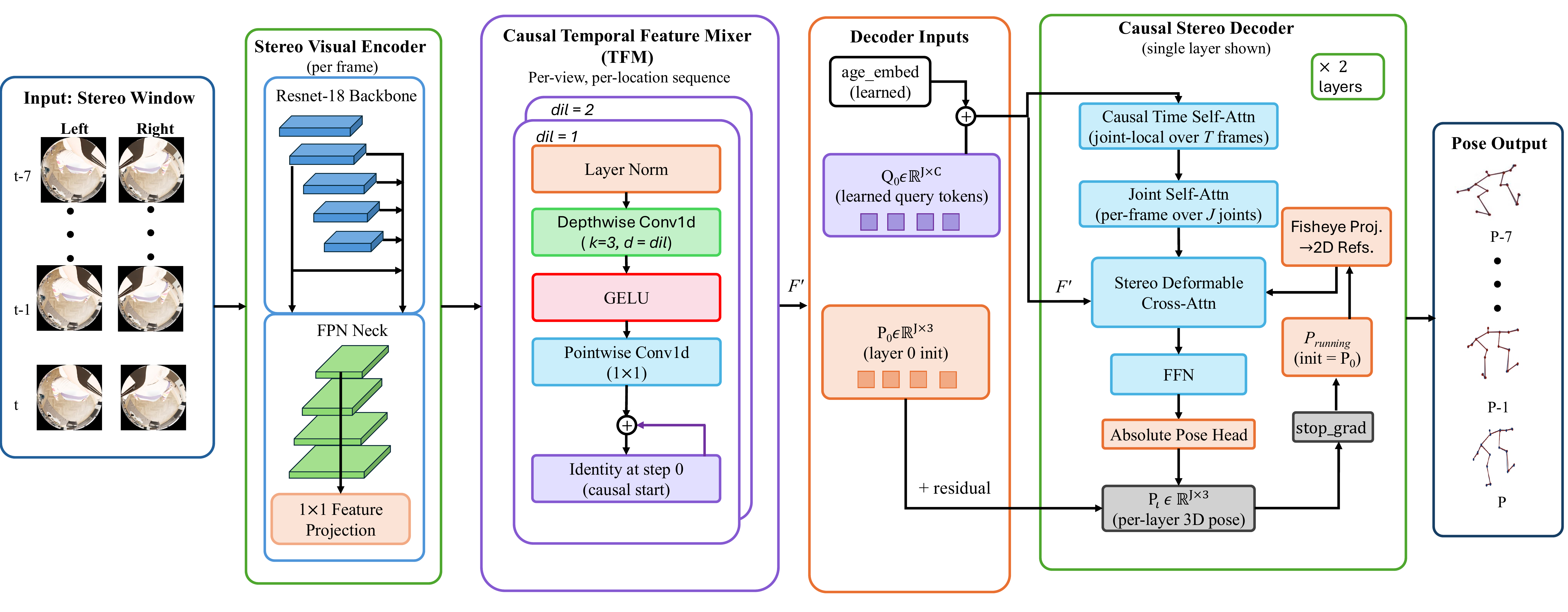}
\vspace{2mm}
\caption{\textbf{Overview of TSR-Ego.} A causal stereo window is encoded by a
shared visual backbone, temporally enriched by the proposed TFM, and decoded by
a single-stage causal stereo decoder. The decoder refines learned 3D joint
queries using temporal self-attention, joint self-attention, and
projection-guided deformable stereo cross-attention. Each layer predicts an
absolute device-relative 3D pose, whose detached projection is used as the
sampling reference for the next layer.}
\label{fig:tgs_pose_architecture}
\end{figure*}

The architecture consists of three stages as depicted in Figure~\ref{fig:tgs_pose_architecture}. First, a stereo visual encoder
extracts dense feature maps for each view and frame. Second, a causal temporal
feature mixer enriches each spatial feature location using only past frames.
Third, a causal stereo decoder iteratively projects the current 3D pose
estimate into each fisheye view, samples image evidence through deformable
cross-attention, and predicts an absolute 3D pose at every decoder layer.

\subsection{Stereo Visual Feature Encoding}

Each input image is encoded independently by a ResNet-18 backbone \cite{resnet} followed by a
feature-pyramid neck \cite{fpn}. The backbone provides features at strides
$\{4,8,16,32\}$, which are fused by the neck into a single stride-4 feature
map. For an input resolution of $256\times256$, this produces a
$64\times64$ feature map. A $1\times1$ convolution projects the neck output to
the decoder hidden dimension $C=128$:
\[
F_{\tau}^{v} = \phi(I_{\tau}^{v}) \in \mathbb{R}^{C\times H\times W}
\]
Stacking all frames and views gives
\[
F \in \mathbb{R}^{B\times T\times V\times C\times H\times W}
\]

We use a dense feature map rather than a global image descriptor because
egocentric pose estimation depends strongly on local geometric evidence. Hands,
feet, and limbs often appear near the image boundary under severe fisheye
distortion, and their visibility varies between the two cameras. Preserving a
stride-4 spatial grid allows later deformable attention to sample local
features around the physically projected joint location instead of forcing all
visual evidence through a global bottleneck.

The encoder is initialized from heatmap pretraining. This gives the backbone an
explicit localization prior before the 3D decoder is trained. We then fine-tune
the encoder jointly with the full model, allowing the features to adapt from
2D heatmap localization to stereo-temporal 3D pose estimation.

\subsection{Causal Stereo-Temporal Feature Mixing}

A central design choice is to inject temporal information into the dense
feature maps before decoding. Given
$F\in\mathbb{R}^{B\times T\times V\times C\times H\times W}$, we treat each
fixed sample, view, and spatial location $(b,v,y,x)$ as a temporal sequence:
\[
F_{b,:,v,:,y,x}\in\mathbb{R}^{T\times C}
\]
We apply a stack of causal depthwise-separable 1D convolutional blocks along
the time axis. For the $m$-th block,
\[
F^{(m+1)} =
F^{(m)} +
\mathrm{PWConv}
\Big(
\mathrm{GELU}
\big(
\mathrm{DWConv}^{\leftarrow}_{k,d}
(\mathrm{LN}_{C}(F^{(m)}))
\big)
\Big),
\]
where $F^{(0)}=F$, $\tilde{F}=F^{(M)}$, $\mathrm{DWConv}^{\leftarrow}_{k,d}$
denotes left-padded causal depthwise convolution over time, $\mathrm{PWConv}$
is a pointwise channel-mixing convolution, and $\mathrm{LN}_{C}$ applies
LayerNorm over channels at each time step.
This module is intentionally temporal-only. It never mixes neighboring spatial
locations. Therefore, the output feature at coordinate $(y,x)$ remains aligned
with the same image coordinate. This is important because the following
deformable stereo attention uses projected 3D joints as geometric reference
points. If temporal modeling were to blur or pool spatial coordinates, the
decoder would sample from feature maps whose spatial positions no longer
correspond cleanly to image locations.

The temporal mixer serves several purposes. It allows visual evidence
from previous frames to support the current frame before the model commits to
joint-level queries. This is useful under self-occlusion, motion blur, extreme
fisheye distortion, and view-dependent limb visibility. In addition, pixel-level
temporal mixing avoids an early query bottleneck. At the beginning of decoding,
the joint queries are learned parameters and have not yet interacted with the
current image evidence. If temporal reasoning were performed only on these
queries, the model would be propagating relatively abstract tokens. By mixing
dense features first, each future cross-attention operation samples from a
feature map that already carries causal temporal context. 

Causality is enforced by left padding only. Thus, the representation at frame
$\tau$ depends on frames $\leq \tau$ and never on future frames. This makes the
model suitable for online inference. We also avoid BatchNorm or GroupNorm in
this module because normalization statistics over the temporal dimension could
leak future information. Per-frame channel LayerNorm preserves causality.

In our default setting, we use two temporal convolution blocks with kernel size
$3$ and dilations $1$ and $2$, giving a receptive field of seven frames. The
pointwise convolution in each block is zero-initialized, so the entire temporal
mixer starts as an identity mapping. This is important for stable optimization:
the model initially behaves like a strong per-frame heatmap-pretrained encoder,
and temporal corrections are learned gradually.

\subsection{Learned Joint Queries and 3D Reference Initialization}

The decoder operates on one token per joint per frame. We initialize the query
content with a learned joint embedding $Q^{0}\in\mathbb{R}^{J\times C}$
(truncated-normal, $\sigma{=}0.02$) and pair it with a learnable initial 3D
reference pose
\[
  P^{0}\in\mathbb{R}^{J\times 3},
\]
which is zero-initialized. Both tensors are broadcast over the batch and
temporal dimensions: the same learned prior seeds every frame, and there is no
separately predicted per-frame initial pose. $Q^{0}$ and $P^{0}$ are optimized
jointly with the rest of the model.

This initialization provides a compact joint-specific prior while letting the
pose estimate be shaped by visual evidence through the decoder. At each decoder
layer, the current 3D reference is projected into each fisheye view to obtain
2D sampling locations for deformable stereo cross-attention; the sampled image
features then update the joint tokens, and a per-layer head regresses an
absolute 3D pose that becomes the reference for the next layer. The sampling
locations are therefore not fixed but evolve with the decoder's own pose
prediction, yielding a geometry-aware refinement process.

The zero initialization of $P^{0}$ is deliberate. Combined with the
zero-initialized final linear of every pose head, it guarantees that at the
first training step every layer predicts exactly $P^{0}$, so training begins
from a clean identity-refinement state and the decoder learns corrections on
top of the learned prior rather than fighting an arbitrary starting pose.

\subsection{Causal Stereo Decoder}

The decoder contains $L$ repeated layers. Each layer performs four operations:
causal temporal self-attention, joint self-attention, fisheye stereo
deformable cross-attention, and feed-forward refinement. After each layer, a
pose head predicts a full absolute 3D pose.

\paragraph{Causal temporal self-attention.}
For each joint independently, we apply self-attention over the $T$ query tokens
corresponding to that joint across time. A strict upper-triangular causal mask
ensures that frame $\tau$ attends only to frames $\leq\tau$:
\[
Q_{\tau,j} \leftarrow
\mathrm{TimeAttn}(Q_{\leq \tau,j}).
\]
We add a learned age embedding before attention, where the current frame has
age $0$, the previous frame has age $1$, and so on.

This component models joint-specific motion history. It is especially useful
for temporally ambiguous observations: a partially visible hand or foot in the
current frame can be disambiguated using its recent trajectory. The age
embedding tells the model how old each token is relative to the target frame,
which is more appropriate than absolute frame indexing for sliding-window
inference. The causal mask preserves online deployment and prevents training
from using information unavailable at test time.

\paragraph{Joint self-attention.}
After temporal reasoning, each frame performs self-attention over the $J$ body
joints:
\[
Q_{\tau} \leftarrow \mathrm{JointAttn}(Q_{\tau,1:J})
\]
This allows the model to reason about body structure. Egocentric images often
contain incomplete local evidence: one arm may be visible in only one camera,
a leg may be truncated, or the torso may be poorly observed because the camera
is head-mounted. Joint self-attention lets confident joints inform uncertain
ones through learned kinematic and pose-context relationships. Unlike a fixed
kinematic tree, attention does not impose a hard skeleton topology; it can
learn dataset-specific dependencies between distant joints when they are
useful.

\paragraph{Fisheye stereo deformable cross-attention.}
The decoder then injects image evidence. For each layer, the current running
3D pose estimate is projected into every fisheye camera:
\[
r_{\tau,j}^{v} =
\Pi_v(P_{\tau,j} + O_{\tau}^{v})
\]
where $P_{\tau,j}$ is the root-relative 3D joint position, $O_{\tau}^{v}$ is
the root location in camera coordinates for view $v$, and $\Pi_v$ is the
calibrated fisheye projection model. The projected 2D point
$r_{\tau,j}^{v}$ is used as the reference point for deformable attention.

For each view, multi-scale deformable attention samples a small set of learned
offset locations around the projected reference point:
\[
Z_{\tau,j}^{v}
=
\mathrm{DeformAttn}(Q_{\tau,j}, \tilde{F}_{\tau}^{v}, r_{\tau,j}^{v})
\]
Although we use a single feature level, deformable attention remains valuable
because it learns where to sample around the projected joint. This handles
projection noise, calibration imperfections, fisheye distortion, and errors in
the current pose estimate. Instead of attending to all image
tokens, the model focuses computation near geometrically plausible locations.

The attention is applied independently to each view. The resulting per-view
features are concatenated and fused with a linear projection:
\[
Z_{\tau,j} = W_f[Z_{\tau,j}^{1}; \ldots ; Z_{\tau,j}^{V}]
\]
If a projected reference point falls outside the field of view, that view's
contribution is masked to zero. This is important in egocentric stereo because
joints frequently appear in only one camera. Masking prevents invalid samples
from contaminating the fused representation while still allowing the other
view to provide evidence.

This module is the main geometry-aware component of the model. The learned
queries decide what visual evidence is needed, while the fisheye projection
constrains where the model should look. This combination is more efficient and
more stable than unconstrained global cross-attention, particularly for
high-distortion egocentric views.

\paragraph{Feed-forward refinement.}
After stereo cross-attention, a two-layer GELU feed-forward network refines the
query representation with a residual connection and LayerNorm. The FFN
increases the layer's capacity beyond attention-based mixing. Attention
aggregates information across time, joints, and image features, while the FFN
performs per-token nonlinear transformation. Empirically, this component is
important because the query must encode both semantic joint identity and
metric 3D pose evidence after cross-attention.

\subsection{Layer-wise Full-Pose Prediction}

Each decoder layer has its own pose head. Given the refined query $Q^{\ell}$,
layer $\ell$ predicts
\[
\hat{P}^{\ell} = P^{0} + h_{\ell}(\mathrm{LN}(Q^{\ell})),
\]
where $h_{\ell}$ is a two-layer MLP and $\hat{P}^{\ell}$ is a complete
root-relative pose in the device frame. Every layer offsets from the same
shared initialization $P^{0}$, not from the previous layer
($\hat{P}^{\ell}\neq\hat{P}^{\ell-1}+\Delta$), so layers are decoupled, errors
do not accumulate, and each layer is interpretable as a full pose estimate.
Each head's final linear is zero-initialized, so all layers start at exactly
$P^{0}$, giving a stable starting point and avoiding random 3D predictions that
project to arbitrary image regions.

The layer-$\ell$ prediction provides the projection reference points for layer
$\ell+1$. We detach it before projection, so sampling gradients do not flow
through the previous head, while the head stays supervised by the MPJPE loss.
Each layer thus learns to predict an accurate pose directly, and later layers
use it as a geometric guide for sampling.

\subsection{Training Objective}

We supervise the final pose prediction and all intermediate decoder predictions
with MPJPE. Let $P^{\star}$ denote the ground-truth pose sequence. The loss is
\[
\mathcal{L}
=
\lambda_{\mathrm{final}}
\mathrm{MPJPE}(\hat{P}^{L}, P^{\star})
+
\lambda_{\mathrm{aux}}
\sum_{\ell=1}^{L-1}
\mathrm{MPJPE}(\hat{P}^{\ell}, P^{\star})
\]
We use $\lambda_{\mathrm{final}}=1.0$ and
$\lambda_{\mathrm{aux}}=0.5$. Deep supervision encourages early decoder layers
to produce meaningful pose estimates, which improves the quality of the projection references used by subsequent layers and stabilizes iterative stereo refinement.

%% file: sec/5_Experiment.tex
\section{Experiments}
\label{sec:experiments}

\subsection{Datasets}
\label{sec:datasets}

We evaluate on two egocentric stereo fisheye datasets that differ in
domain, scale, and camera calibration.

\paragraph{UnrealEgo2~\cite{3dposeperception}.}
UnrealEgo2 is a large-scale synthetic dataset rendered in Unreal Engine from
motion-capture sequences. Each sample contains synchronized left/right
head-mounted fisheye images at $1024{\times}1024$ resolution, together with
accurate 3D pose annotations. Although the raw annotations contain a much
larger skeleton, we follow our model setting and use a $16$-joint limb subset.
The pelvis is used only as the root for pose normalization and as the per-view
projection origin; it is not itself a predicted joint. Ground-truth poses are
expressed in the head-mounted device frame and centered at the pelvis, i.e.
device-relative coordinates with the pelvis as origin
($P_j = \mathrm{joint}_j^{\text{dev}} - \mathrm{pelvis}^{\text{dev}}$). The
synthetic stereo cameras share the Scaramuzza fisheye calibration used by the
UnrealEgo rendering setup. We follow the official train/validation/test split.

\paragraph{UnrealEgo-RW~\cite{3dposeperception}.}
UnrealEgo-RW is the real-world counterpart, captured with two physically
separate head-mounted fisheye cameras at $872{\times}872$ resolution. Unlike
the synthetic setting, the two cameras have distinct per-view Scaramuzza
calibrations. The dataset provides $16$ limb joints and no pelvis; we therefore
define the root as the mid-hip, the midpoint of \texttt{LeftUpLeg} and
\texttt{RightUpLeg}. Ground-truth poses are expressed in the device frame,
relative to this mid-hip root, and all $16$ joints are supervised. We use the
official train/validation/test split of $402$, $96$, and $93$ sequences.

\subsection{Training Details}
\label{sec:training_details}

We use a ResNet-18 backbone with an FPN neck as the visual encoder. The
encoder is initialized from a dataset-specific heatmap-regression checkpoint
and then optimized jointly with the 3D decoder. The model takes a causal stereo window of $T{=}8$ frames as input. We sample
training windows with stride $4$ and validation/test windows with stride $8$.

All models are trained for $12$ epochs on a single GPU using AdamW
\cite{adamw}. The initial learning rate is $10^{-3}$, weight
decay is $5{\times}10^{-4}$, and the learning rate is warmed up linearly for
the first $200$ iterations. We decay the learning rate by a factor of $0.1$ at
epochs $8$ and $10$. The batch size is $16$. The loss weights are
$w_{\mathrm{final}}=1.0$ for the final decoder prediction and
$w_{\mathrm{aux}}=0.5$ for auxiliary supervision on intermediate decoder
layers.

\subsection{Quantitative Evaluation}
\label{sec:quantitative}

\begin{table}[t]
\centering
\begin{tabular}{lccc}
\toprule
Method & MPJPE$\downarrow$ & PA-MPJPE$\downarrow$ & 3D PCK$\uparrow$ \\
\midrule
EgoGlass~\cite{egoglass} & 79.64 & 58.22 & 88.50 \\
UnrealEgo~\cite{unrealego} & 72.80 & 52.88 & 91.32 \\
Akada et al.~\cite{3dposeperception} & 30.53 & 26.72 & 97.22 \\
EgoPoseFormer~\cite{egoposeformer} & 23.12 & 21.69 & 97.69 \\
\textbf{Ours} & \textbf{22.36} & \textbf{21.23} & \textbf{98.36} \\
\bottomrule
\end{tabular}
\vspace{2mm}
\caption{Quantitative results for device-relative 3D pose estimation on
UnrealEgo2. MPJPE and PA-MPJPE are reported in millimeters.}
\label{tab:unrealego2_results}
\end{table}

\begin{table}[t]
\centering
\begin{tabular}{lccc}
\toprule
Method & MPJPE$\downarrow$ & PA-MPJPE$\downarrow$ & 3D PCK$\uparrow$\\
\midrule
EgoGlass~\cite{egoglass} & 117.57 & 88.01 & 73.12 \\
UnrealEgo~\cite{unrealego} & 122.64 & 86.55 & 72.51 \\
Akada et al.~\cite{3dposeperception} & 104.14 & 82.18 & 80.20 \\
EgoPoseFormer~\cite{egoposeformer} & 74.31 & 61.79 & 82.84 \\
\textbf{Ours} & \textbf{65.58} & \textbf{56.03} & \textbf{90.66} \\
\bottomrule
\end{tabular}
\vspace{2mm}
\caption{Quantitative results for device-relative 3D pose estimation on
UnrealEgo-RW. MPJPE and PA-MPJPE are reported in millimeters.}
\label{tab:unrealego_rw_results}
\end{table}

We evaluate device-relative 3D pose estimation on both synthetic and real-world
egocentric benchmarks. Tables~\ref{tab:unrealego2_results}
and~\ref{tab:unrealego_rw_results} report results on UnrealEgo2 and
UnrealEgo-RW, respectively, using MPJPE, PA-MPJPE, and 3D PCK.
Since EgoPoseFormer \cite{egoposeformer} does not report results on UnrealEgo2 or UnrealEgo-RW, we train the official implementation using the same train/validation/test splits, input resolution, joint subset, root normalization, and evaluation metrics as TSR-Ego.
On UnrealEgo2, our method obtains the best performance across all reported
metrics, with $22.36$ mm MPJPE, $21.23$ mm PA-MPJPE, and $98.36\%$ 3D PCK.
Compared with EgoPoseFormer, our model reduces MPJPE by $0.76$ mm and
PA-MPJPE by $0.46$ mm. The margin is moderate because UnrealEgo2 is a clean
synthetic benchmark where stereo observations are relatively reliable and the
strongest transformer-based methods already operate near saturation. Still,
the consistent improvement in both MPJPE and PA-MPJPE indicates that the
proposed temporal feature aggregation improves pose estimation without relying
only on better rigid alignment.

The gain is larger on UnrealEgo-RW, which contains real image noise, motion
blur, illumination variation, calibration imperfections, and frequent
self-occlusion. Our method achieves $65.58$ mm MPJPE and $56.03$ mm PA-MPJPE,
improving over EgoPoseFormer by $8.73$ mm and $5.76$ mm, respectively. This
larger gap suggests that the proposed causal temporal modeling is most useful
when single-frame stereo evidence is unreliable. By enriching pixel-level
stereo features with information from previous frames before deformable
cross-attention, the decoder can sample from temporally stabilized features
while preserving the spatial structure required for accurate stereo reasoning. Our
method also reaches $90.66\%$ 3D PCK, outperforming EgoPoseFormer~\cite{egoposeformer}
with available PCK by nearly $8\%$.

\begin{figure}[t]
    \centering
\includegraphics[width=0.9\linewidth, trim=0 0 60 0, clip]{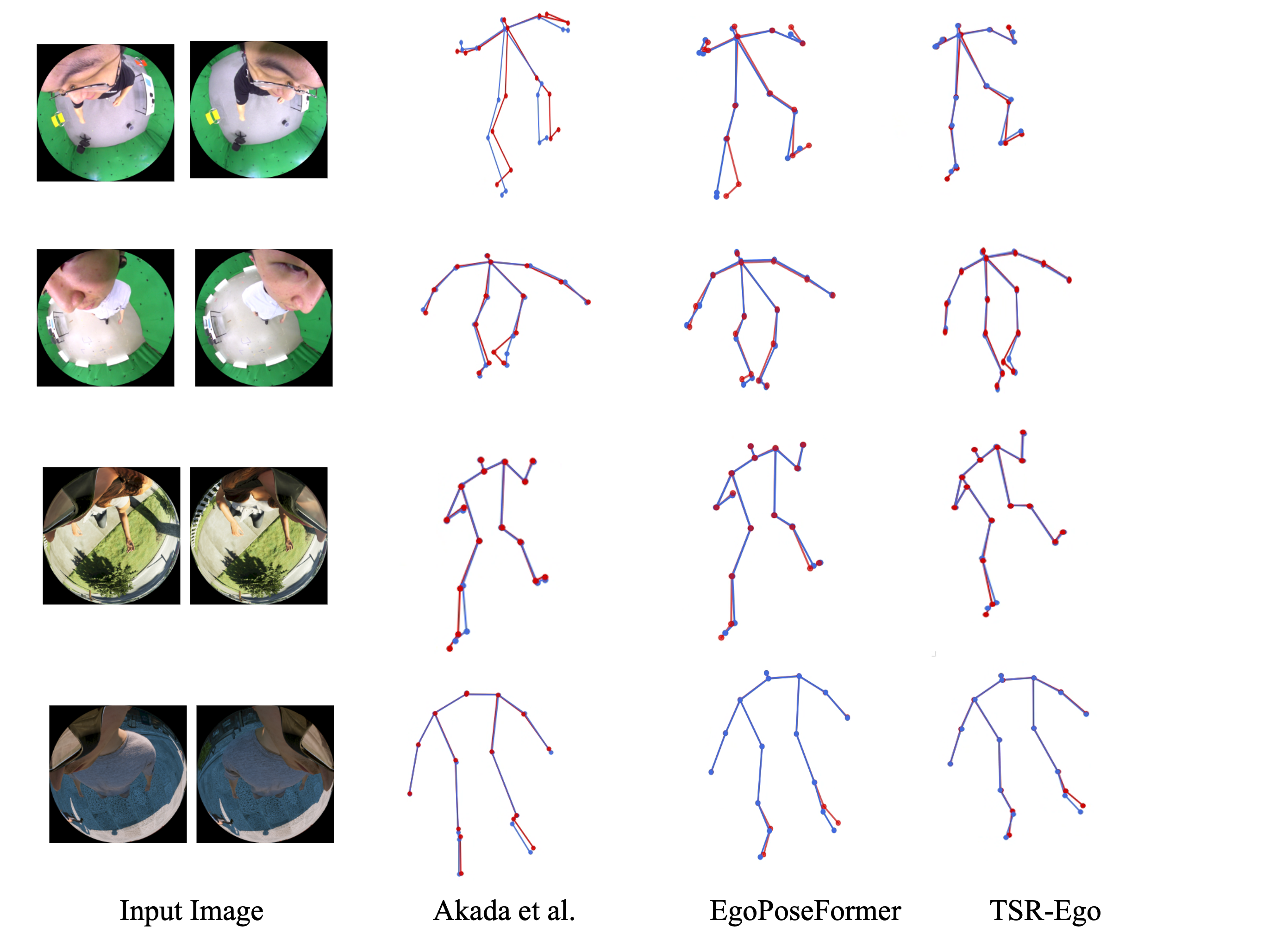}

    \caption{Qualitative comparison on UnrealEgo-RW (rows 1--2) and
    UnrealEgo2 (rows 3--4). Each row shows stereo fisheye input (left)
    with predicted 3D poses from Akada~et~al.~\cite{3dposeperception},
    EgoPoseFormer~\cite{egoposeformer}, and our TSR-Ego.Red skeletons denote ground truth and blue skeletons denote predictions.}
    \label{fig:qualitative}
\end{figure}

\subsection{Qualitative Results}

Figure~\ref{fig:qualitative} compares predicted 3D poses from
Akada~et~al.~\cite{3dposeperception}, EgoPoseFormer~\cite{egoposeformer}, and TSR-Ego
on four sequences. On the real-world sequences (rows 1--2), which involve challenging
lighting and fine-grained upper-body motion, Akada~et~al.\ shows
notable limb orientation errors while EgoPoseFormer exhibits drift in
the leg joints. TSR-Ego recovers both arm and leg configurations more
accurately, benefiting from causal temporal feature mixing over the
8-frame window. On the synthetic sequences (rows 3--4), which feature
full-body dynamic motion, TSR-Ego produces the tightest alignment
with ground truth, with reduced errors in the extremities. These results
demonstrate that the proposed temporal pixel-level enrichment and
joint-local causal decoder generalise across both real-world and
synthetic egocentric capture conditions.

\subsection{Ablation Studies}
\label{subsec:ablation}
We conduct ablation studies on UnrealEgo-RW to evaluate the contributions of the proposed decoder components, temporal window length, decoder depth, stereo input, and temporal feature mixer design.

\subsubsection{Component Ablation}
\label{subsec:component_ablation}

In Table \ref{tab:ablation}, we ablate the main components of our proposed model on
UnrealEgo-RW. Each variant is trained from scratch under the same setting as the full model.

\begin{table}[t]
\centering
\begin{tabular}{cccccc}
\toprule
TFM & Age & Joint & FFN & MPJPE$\downarrow$ & PA-MPJPE$\downarrow$ \\
\midrule
\checkmark & \checkmark & \checkmark & \checkmark
& \textbf{65.58} & \textbf{56.03} \\
           & \checkmark & \checkmark & \checkmark
& 68.99 & 58.91 \\
\checkmark &             & \checkmark & \checkmark
& 67.11 & 58.31 \\
\checkmark & \checkmark &             & \checkmark
& 67.57 & 58.62 \\
\checkmark & \checkmark & \checkmark &
& 69.78 & 59.74 \\
\bottomrule
\end{tabular}
\vspace{2mm}
\caption{Component-wise ablation on UnrealEgo-RW. TFM denotes the causal
temporal feature mixer, Age denotes the temporal age embedding, Joint denotes
joint self-attention, and FFN denotes the feed-forward network.}
\label{tab:ablation}
\end{table}
The full model achieves the best performance, showing that all four components
contribute to the final accuracy. Removing the FFN causes the largest drop,
increasing MPJPE from $65.58$ mm to $69.78$ mm. This suggests that the
per-query non-linear transformation after attention is important for converting
the sampled stereo features into accurate 3D joint estimates.

The temporal feature mixer is the second most influential component. Without
TFM module, MPJPE increases by $3.41$ mm and PA-MPJPE by $2.88$ mm. This supports our design motivation: temporal information is most effective
when injected into the stereo feature maps before deformable cross-attention,
so that the decoder samples from temporally enriched visual evidence rather
than relying only on query-level temporal reasoning.

Joint self-attention and age embedding give smaller but consistent gains.
Removing joint self-attention degrades MPJPE by $1.99$ mm, indicating that
explicit cross-joint interaction helps preserve body structure. Removing the
age embedding increases MPJPE by $1.53$ mm, showing that the decoder benefits
from knowing the relative temporal position of each frame in the causal window.
\subsubsection{Temporal Window Length}

We further study the effect of the causal temporal window length $T$ on
UnrealEgo-RW. As shown in Table~\ref{tab:ablation_temporal_length}, increasing
the temporal context consistently improves MPJPE, with the best result obtained
at $T=8$. Compared with the single-frame setting ($T=1$), using eight frames
reduces MPJPE by $4.30$ mm. This confirms that past frames provide useful
visual evidence for resolving ambiguous egocentric observations like occlusion and motion blur.

The improvement saturates as $T$ increases. PA-MPJPE remains within a narrow
range across $T=2$ to $T=8$, suggesting that longer temporal context mainly
helps device-relative localization rather than improving the aligned body
shape. We therefore use $T=8$, which provides the best
MPJPE while keeping the temporal window causal and compact.

\begin{table*}[t]
\centering
\setlength{\tabcolsep}{4pt}
\footnotesize

\begin{minipage}[t]{0.46\textwidth}
\centering
\begin{tabular}{@{}ccc@{}}
\toprule
Length $T$ & MPJPE$\downarrow$ & PA-MPJPE$\downarrow$ \\
\midrule
1 & 69.88 & 57.97 \\
2 & 68.42 & 56.97 \\
4 & 67.97 & 57.18 \\
6 & 66.92 & 57.02 \\
8 & \textbf{65.58} & \textbf{56.03} \\
\bottomrule
\end{tabular}
\vspace{2mm}
\captionof{table}{Ablation on the causal temporal window length $T$ on
UnrealEgo-RW. All metrics are reported in millimeters.}
\label{tab:ablation_temporal_length}
\end{minipage}
\hfill
\begin{minipage}[t]{0.46\textwidth}
\centering
\begin{tabular}{@{}ccc@{}}
\toprule
Depth & MPJPE$\downarrow$ & PA-MPJPE$\downarrow$ \\
\midrule
1 & 74.40 & 63.27 \\
2 & 65.58 & 56.03 \\
3 & 63.76 & 55.62 \\
4 & \textbf{62.80} & \textbf{55.30} \\
\bottomrule
\end{tabular}
\vspace{2mm}
\captionof{table}{Ablation on decoder depth on UnrealEgo-RW. All metrics are
reported in millimeters.}
\label{tab:ablation_decoder_depth}
\end{minipage}

\end{table*}

\subsubsection{Decoder Depth}

We also vary the number of decoder layers to analyze the effect of iterative
stereo feature aggregation. As shown in Table~\ref{tab:ablation_decoder_depth},
a single decoder layer performs substantially worse, indicating that one round
of joint reasoning and deformable stereo attention is insufficient for accurate
3D pose recovery. Increasing the depth from one to two layers reduces MPJPE by
$8.82$ mm and PA-MPJPE by $7.24$ mm, showing the importance of iterative pose
refinement.

Using three or four layers further improves accuracy, reaching $62.80$ mm
MPJPE and $55.30$ mm PA-MPJPE at depth four. However, the gains beyond two
layers are comparatively smaller than the improvement from one to two layers.
In our final model, we use two decoder layers as a trade-off between accuracy and
computational cost. This setting already captures most of
the benefit of iterative decoding while keeping the architecture lightweight
for causal egocentric pose estimation.

\subsubsection{Monocular vs. Stereo Input}

Table~\ref{tab:ablation_stereo} evaluates the effect of stereo input. Both
monocular variants are substantially worse than the stereo model, with the
left-only and right-only settings obtaining $75.35$ mm and $71.94$ mm MPJPE,
respectively. Using both views reduces MPJPE to $65.58$ mm and PA-MPJPE to
$56.03$ mm. This confirms that binocular observations provide important depth
cues for device-relative 3D pose estimation, especially for absolute joint
localization in the headset coordinate frame.

\begin{table*}[t]
\centering
\setlength{\tabcolsep}{4pt}
\footnotesize

\begin{minipage}[t]{0.46\textwidth}
\centering
\begin{tabular}{@{}lcc@{}}
\toprule
Input & MPJPE$\downarrow$ & PA-MPJPE$\downarrow$ \\
\midrule
Left only  & 75.35 & 63.66 \\
Right only & 71.94 & 61.36 \\
Stereo     & \textbf{65.58} & \textbf{56.03} \\
\bottomrule
\end{tabular}
\vspace{2mm}
\captionof{table}{Ablation on monocular and stereo input on UnrealEgo-RW.}
\label{tab:ablation_stereo}
\end{minipage}
\hfill
\begin{minipage}[t]{0.46\textwidth}
\centering
\begin{tabular}{@{}lcc@{}}
\toprule
TFM design & MPJPE$\downarrow$ & PA-MPJPE$\downarrow$ \\
\midrule
$L=1,K=3$ & 67.23 & 58.06 \\
$L=2,K=3$ (Ours) & \textbf{65.58} & \textbf{56.03} \\
$L=3,K=3$ & 68.14 & 57.19 \\
$L=2,K=5$ & 65.65 & 56.73 \\
$L=1,K=5$ & 66.44 & 57.35 \\
\bottomrule
\end{tabular}
\vspace{2mm}
\captionof{table}{Ablation on the temporal feature mixer design. $L$ denotes
the number of causal temporal blocks and $K$ denotes the temporal kernel size.}
\label{tab:ablation_tfm_design}
\end{minipage}

\end{table*}

\subsubsection{Temporal Feature Mixer Design}

Table~\ref{tab:ablation_tfm_design} studies the design of the temporal feature
mixer. Here, $L$ is the number of causal temporal convolution blocks and $K$ is
the temporal kernel size. Our final setting uses $L=2,K=3$, which achieves the
best result among the tested designs. Reducing the mixer to one block weakens
temporal aggregation, while increasing it to three blocks does not improve
accuracy, suggesting that overly deep temporal mixing may introduce noisy
history from less relevant frames. Increasing the kernel to $K=5$ performs
competitively, but remains slightly worse than $L=2,K=3$ and adds extra
temporal computation. We therefore use $L=2,K=3$ as a compact and effective
design.

\begin{figure*}[t]
\centering
\begin{minipage}[t]{0.48\textwidth}
\centering
\includegraphics[width=\linewidth]{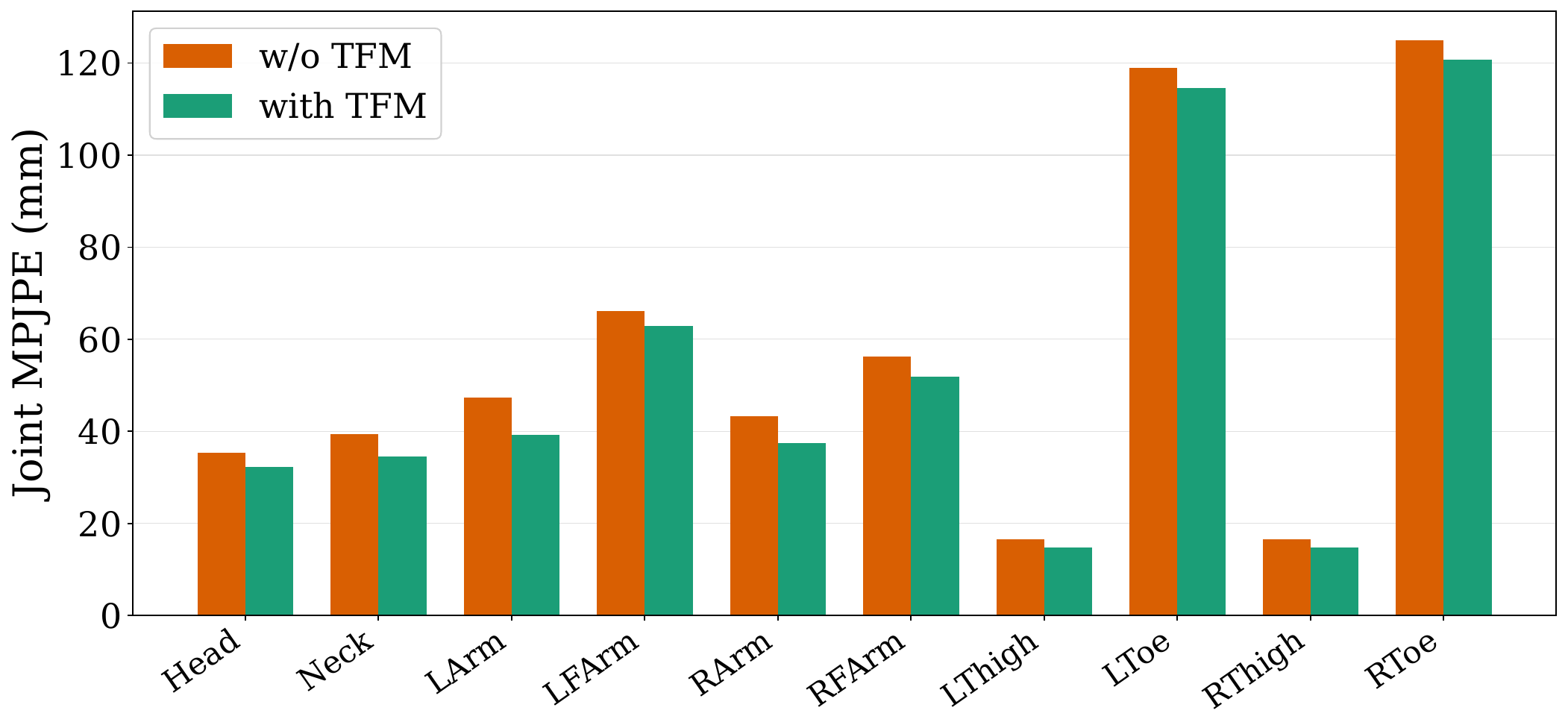}
\vspace{-2mm}
\caption{Joint-wise MPJPE comparison on selected UnrealEgo-RW joints with and
without the temporal feature mixer (TFM).}
\label{fig:per_joint_tfm}
\end{minipage}
\hfill
\begin{minipage}[t]{0.48\textwidth}
\centering
\includegraphics[width=\linewidth]{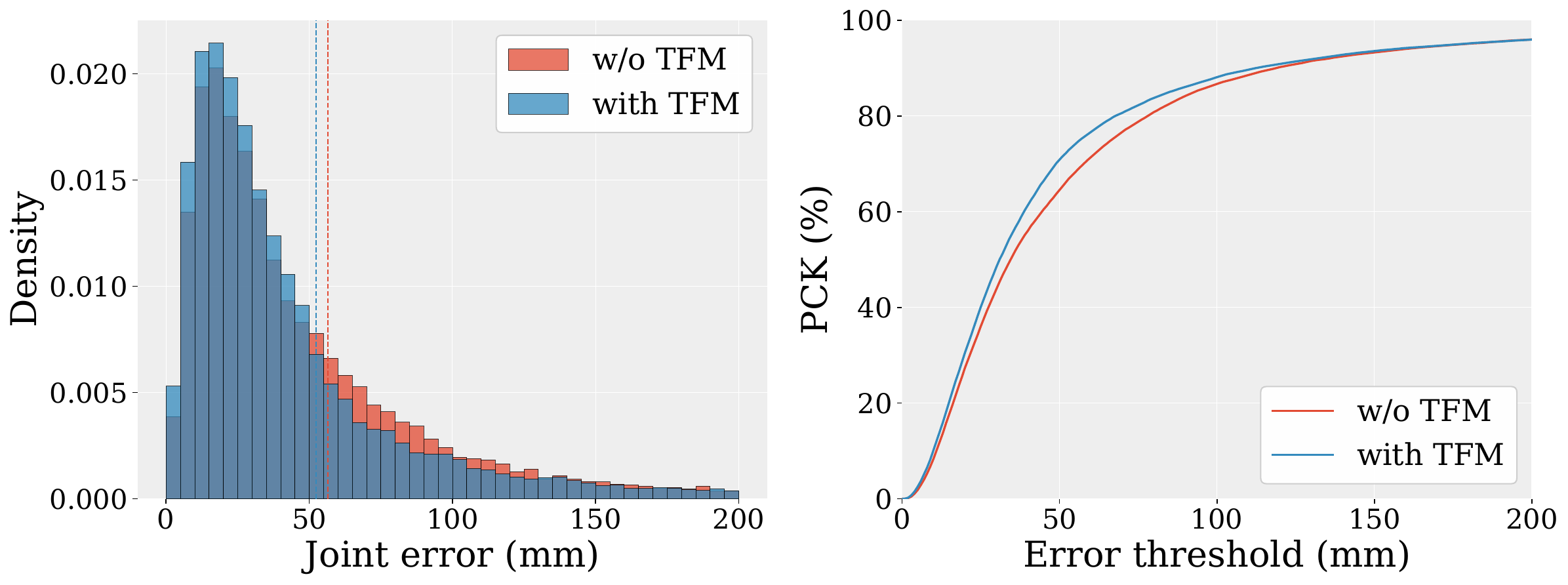}
\vspace{-2mm}
\caption{Joint-error distribution and PCK curve on UnrealEgo-RW with and
without the temporal feature mixer (TFM).}
\label{fig:error_distribution_tfm}
\end{minipage}
\end{figure*}

\paragraph{Error Analysis.}
Figures~\ref{fig:per_joint_tfm} and~\ref{fig:error_distribution_tfm} analyze
the effect of the temporal feature mixer (TFM). TFM reduces mean MPJPE from
$68.99$ mm to $65.58$ mm, with the largest gains on LeftArm ($8.10$ mm),
RightArm ($5.78$ mm), and Neck ($4.95$ mm), which are often affected by
self-occlusion and rapid egocentric motion. Consistent improvements on forearms
and toe bases further show that temporal context benefits both articulated body
structure and distal joints.

The error distribution shifts toward lower errors, and the PCK curve remains
higher across most thresholds. The gains are strongest in the low-to-mid error
range, suggesting that TFM mainly reduces moderately difficult localization
errors. At larger thresholds, the curves become closer, indicating that severely
occluded joints remain challenging.

%% file: sec/6_Conclusion.tex
\section{Conclusion}
\label{sec:conclusion}

We introduced TSR-Ego, a single-stage framework for egocentric 3D pose estimation from head-mounted stereo fisheye cameras. TSR-Ego injects causal temporal context directly into dense stereo features before pose decoding, enabling projection-guided deformable attention to sample temporally enriched evidence while preserving spatial alignment. A lightweight decoder then refines learned 3D joint queries through temporal, joint-wise, and stereo cross-attention to predict 3D poses end-to-end.

Experiments on UnrealEgo2 and UnrealEgo-RW show consistent gains over strong egocentric baselines, especially in real-world settings where frame-local stereo cues are often ambiguous. The ablation studies show consistent contributions from temporal feature mixing, feed-forward refinement, age embedding, joint self-attention, and decoder depth, indicating that TSR-Ego benefits from both temporal feature enrichment and iterative query refinement.

TSR-Ego still relies on calibrated and synchronized stereo input, and distal joints remain challenging under severe occlusion or truncation. Future work will explore calibration-robust stereo attention, stronger kinematic and motion priors, and broader real-world egocentric capture scenarios.